\documentclass[10pt,a4paper]{article}
 
\usepackage[T2A]{fontenc}
\usepackage[utf8]{inputenc}
\usepackage[english]{babel}
\usepackage{amsmath}
\usepackage{amsfonts}
\usepackage{amssymb}
\usepackage[style=numeric,backend=bibtex,sorting=none]{biblatex}
\usepackage{algorithm}
\usepackage{algpseudocode}
\usepackage{graphicx}
\usepackage{subfigure}
\usepackage[a4paper]{geometry}

\begin{document}
\title{Neural Network Optimization for Reinforcement Learning Tasks Using Sparse Computations}
\author{
Dmitry Ivanov\\
\textit{rudimiv@gmail.com}\\
Cifrum, Lomonosov Moscow State University \and 
Michail Kiselev\\
\textit{mkiselev@chuvsu.ru}\\
Cifrum, Chuvash State University \and
Denis Larionov\\
Cifrum\\
\textit{denis.larionov@gmail.com} 
}
\date{January 2022}
\maketitle

\begin{abstract}

This article proposes a sparse computation-based method for optimizing neural networks for reinforcement learning (RL) tasks. This method combines two ideas: neural network pruning and taking into account input data correlations; it makes it possible to update neuron states only when changes in them exceed a certain threshold. It significantly reduces the number of multiplications when running neural networks. We tested different RL tasks and achieved 20-150x reduction in the number of multiplications. There were no substantial performance losses; sometimes the performance even improved.
\end{abstract}


\newpage
\section{Intorduction}

Many modern neural networks consume an enormous amount of computing resources. This is due to the long time it takes to train a neural network, large data sets, and large number of parameters in them. Some recent studies  \cite{blalock2020state, hinton_sparse, frankle2018lottery} showed that many weights are excessive and can be removed without loss (or with a small loss) of neural network performance.

On the other hand, input data for some neural network tasks are sequences of highly correlated frames. Video / audio processing and RL tasks are good examples of such tasks. In such tasks, what a neural network saw at step t-1 is very similar to what it sees now at step t. Some studies propose various optimization algorithms for neural networks handling such data \cite{sparnet2020, delta_rnn, oconnor_sigma}. These approaches are based on the idea of asynchronous update of the states of only those neurons that changed significantly compared to the previous step.

Neural network layers can be represented as a combination of matrix-vector multiplication and application of a nonlinear transformation. Therefore, any value of the output vector can be represented as  $y_i = f(w_{i1} * x_1 + ... + w_{in} * x_n)$, which is a superposition of a nonlinear function and Multiply and Accumulate (MAC) operation. A MAC operation is the summation of the results of products of the input vector elements and the respective weights. If at least one operand in any of such multiplications is zero, then such multiplication could be omitted. We will call a multiplication of two numbers  \textit{significant} if both its operands are not equal to zero.

In this study, we focused on RL tasks and applied a combination of the two abovementioned approaches to optimize them. The first approach yields a 2-8x reduction in the number of significant neural network multiplications; the second one yields a 10-50x reduction. Combined, they yield a 20-150x reduction in the number of significant multiplications without substantial performance losses; sometimes the performance even improved. To the best of our knowledge, a combination of these methods has never been applied to RL tasks.

It is worth noting that brain neurons also work asynchronously and send signals to each other only when necessary. In addition, there are no dense layers in the brain, while they are characteristic of classic neural networks. This suggests that the combination of the proposed methods is biologically inspired.

\section{Methods}

\subsection{Deep Q-Network}

In RL tasks, an agent receives current environment state $s$ as input, after which it selects action $a$, and then goes to new state $s'$ and receives certain reward $r$. The agent's goal is to maximize the sum of rewards.

More strictly, the environment is formalized as a Markov decision process. A Markov decision process (MDP) is a tuple $(S, A, P, R)$, where $S$ is a set of possible states, $A$ is a set of possible actions. $P$ is the function describing transition between states; $P_{a}(s, s') = Pr(s_{t+1} = s' | s_t = s, a_t = a)$, i.e. the probability to get into state $s'$ at the next step when selecting action $a$ in state $s$. $R = R_{a}(s,s')$ is the function describing receiving rewards; it determines how big is a reward that an agent will receive when transitioning from state $s$ to state $s'$ by selecting action $a$.

A strategy an agent uses to selects its actions $a$ depending on state $s$ is called a policy and is usually denoted by letter $\pi_\theta$ Where $\theta$ denotes policy parameters.

One way to solve an RL task are Q-function-based approaches. A Q-function has two parameters – $s$ and $a$. $Q_{\pi_\theta}(s,a)$ determines what reward agent ${\pi_\theta}$ will receive if it performs action $a$ from state $s$ and then follows policy ${\pi_\theta}$. If complete information about an environment is available, the exact value of a Q-function can be calculated. However, the knowledge of the world is usually incomplete and the number of possible states is enormous. Therefore, a Q-function has to be approximated using neural networks.

This approach was successfully demonstrated by DeepMind in \cite{mnih2015human}. The neural network architecture presented in this study is called DQN (Deep Q-Network). For the purposes of this study, it is the main architecture for optimization.

This neural network is as follows:

A 84x84x4 matrix that is received from the environment is input to the neural network. It is input for the first convolutional layer consisting of 32 8x8 filters with strides equal to 4 and ReLU activations. The second layer consists of 32 4x4 filters with strides equal to 2 and ReLU activation. The third layer consists of 64 3x3 filters with strides equal to 2 and ReLU activation. They are followed by a dense layer with 512 neurons with ReLU activations. At the output there is another dense layer with a number of neurons equal to the number of actions in a video game. Depending on the video game, the number of actions $n$ may vary from 4 to 18.

The neural network structure is shown in table:

\begin{table}
\begin{tabular}{ |c|c|c| } 
\hline
Layer & Input Shape & Param  \\
\hline
Conv2d-1 (8x8, stride=4) & [4, 84, 84] & 8,224 \\
Conv2d-2 (4x4, stride=2) & [32, 20, 20] & 32,832 \\
Conv2d-4 (3x3, stride=1) & [64, 9, 9] & 36,928 \\
Flatten & [64, 7, 7]  & 0 \\
Dense-1 (3136, 512) & [3136]  & 1,606,144 \\
Dense-2 (512, $n_{output}$) & [512] & $512 * n_{output} + n_{output}$ \\
\hline
\end{tabular}
\caption{Deep Q-Network structure}
\label{table:structure}
\end{table}


\begin{figure}
  \includegraphics[width=\linewidth]{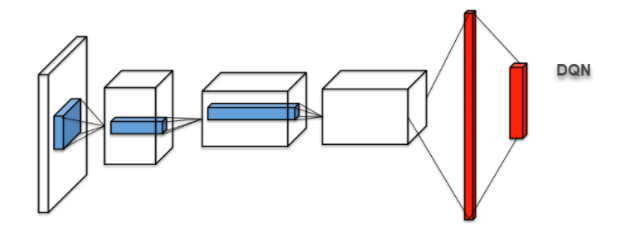}
  \caption{DQN architecture. DQN consists of three convolutional layers and two dense layers. This architecture is suitable for all the video games (the number of outputs at the last layer is the only value that changes)}
  \label{fig:DQN}
\end{figure}

Double q-learning \cite{double_q_learning} was also used for the training.

We experimented and optimized within the following RL environments: Breakout, SpaceInvaders.

\begin{figure}
\subfigure[Breakout]{\includegraphics[width=0.16\textwidth]{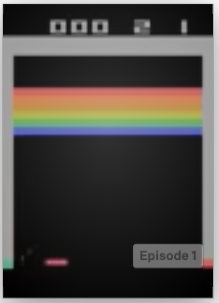}}
\subfigure[Spaceinvaders]{\includegraphics[width=0.16\textwidth]{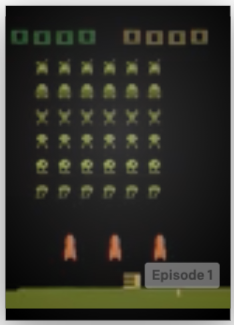}}
\subfigure[BattleZone]{\includegraphics[width=0.16\textwidth]{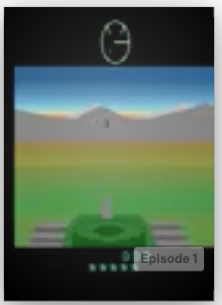}}
\subfigure[Robotank]{\includegraphics[width=0.16\textwidth]{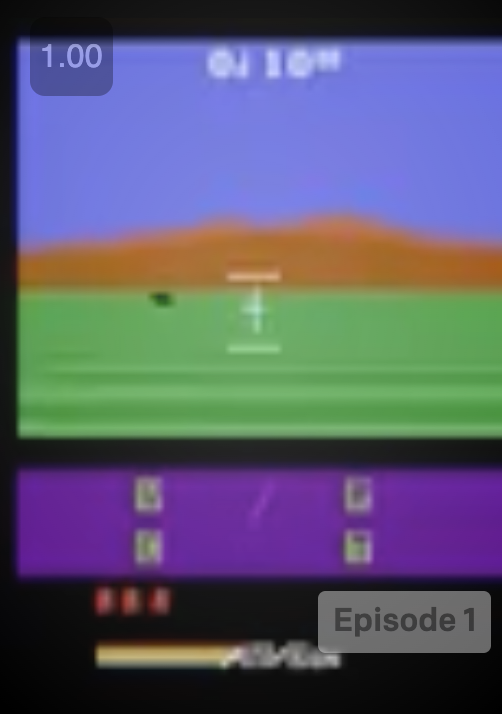}}
\subfigure[Enduro]{\includegraphics[width=0.16\textwidth]{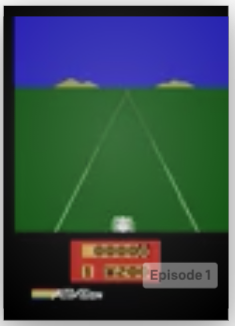}}
\subfigure[Freeway]{\includegraphics[width=0.16\textwidth]{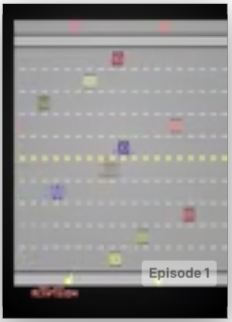}}

\caption{RL environments. For example, in BreakOut, an agent has to hit as many bricks as possible by hitting the ball. In Spacelnvaders, an agent has to eliminate all alien spaceships.}

\label{fig:Atari}
\end{figure}

Image \ref{fig:Atari} shows frames from these video games.

\subsection{Pruning, Lottery Ticket}

Pruning – i.e. removal of unnecessary neural network weights – is one of the ways to structural sparsity within a neural network.

Neural network pruning is not a novel idea; it was conceived back in the 90s   \cite{lecun1990optimal, hassibi1993surgeon}. There are various strategies to identify redundant connections in a neural network (by absolute value, by analyzing the Hessian, etc.).

For this study, we used weight pruning by absolute values: the closer a weight is to zero, the less significant such weight is.


The authors of \cite{frankle2018lottery} have discovered the following interesting fact: if at any stage of neural network training weight values are pruned by absolute value and the remaining weights are reset to the values they had before the training, and then neural network training is resumed , the training of such neural network will go well. However, this does not happen if we set the remaining weights to random (not initial) values. Sparse neural network performance after such training can turn out to be better than unpruned neural network performance.

However, in practice \cite{lotteryRL1, vischer2021lottery, frankle2018lottery} neural network iterative magnitude pruning by absolute value is usually used: neural networks are trained, then a small percentage of their weights (10-20 \%) are pruned, then the remaining weights are reset to initial values, then this procedure is repeated $n$ times. Thus, for the pruning rate $r$, the fraction of pruned weights will be calculated according to the following formula:

\begin{equation}
\frac{pruned\_weights}{total\_weights} = 1 - (1-r)^i,
\end{equation}

\begin{table}
\begin{tabular}{ |c|c|c|c|c|c|c|c|c|c|c| } 
\hline
i & 0 & 1 & 2 & 3 & 4 & 5 & 6 & 7 & 8 & 9\\
\hline
$\frac{pruned\_weights}{total\_weights}$  & 0.000 & 0.200 & 0.36 & 0.488 & 0.590 & 0.672 & 0.730 & 0.790 & 0.832 & 0.866 \\
\hline
\end{tabular}
\caption{Pruned weights fraction dependence on iteration number}
\label{table:structure}
\end{table}


where $i$ is the pruning iteration, $r$ is the percentage of pruned weights at every iteration (pruning rate).

Authors of \cite{lotteryRL1, vischer2021lottery} showed that this phenomenon is observed in RL tasks as well. Authors of \cite{vischer2021lottery} explored this phenomenon in detail for both DQN and PPO algorithm. However, they studied it on small MinAtar subtasks \cite{minatar} (MinAtar is a simplified version of classic AtariGames).


\subsection{DeltaNetwork}

The main idea behind this approachn\cite{yousefzadeh2019asynchronous, sparnet2020} ] is to use temporal sparsity when working with sequential data.

The output values of the neural network layer number  $k+1$ can be written as:

\begin{equation}
o^{k} = W^{k} x^{k} + b^{k}
\end{equation}

\begin{equation}
x^{k+1} = f(o^{k})
\end{equation}


where $x^{k+1} \in R^n$ is the output of $k+1$ neural network layer, $x^k \in R^n$ is the input of $k+1$ neural network layer (output of $k$ layer), $W^{k} \in R^{nxn}$ is the weight matrix, and $b^{k} \in R^{n}$ is the bias. In a conventional neural network, for every new input vector $x^{k}(t)$ in the moment of time t a total recomputation of output value $x^{k+1}(t)$ is required, which will require $n^2 + n$ multiplications. However, the following should be noted:

\begin{equation} \label{delta_eq:1}
\Delta x^k(t) = x^k(t) - x^k(t-1)
\end{equation}

\begin{equation} \label{delta_eq:2}
o^{k}(t) = W^{k} \Delta x^{k}(t) + o^{k}(t-1)
\end{equation}

\begin{equation} \label{delta_eq:3}
\Delta x^{k+1}(t) = f(o^{k}(t)) - f(o^{k}(t-1))
\end{equation}

Thus, it is possible to recompute layer output values at the moment of time t using equations  \ref{delta_eq:1}, \ref{delta_eq:2}, \ref{delta_eq:3} using layer input changes that occur relative to the state at the moment of time  $t-1$.

However, this remark does not lead to neural network optimization by itself. But we can introduce threshold T for output value changes  $\Delta x^{k}(t)$ such that recomputation of succeeding neurons is started only when an output value exceeds this threshold. 

Authors of \cite{sparnet2020, yousefzadeh2019asynchronous} call this approach "Hysteresis Quantizer". To implement it, it is necessary to introduce an additional variable into each neuron; such variable will be used to record last transmitted value. Thus, the following algorithm will be run on each neuron:


\begin{algorithm}
\caption{Update neuron j at time t}\label{alg:cap}
\begin{algorithmic}

\For{\texttt{each incoming data $\Delta X_{i}(t)$ from ancestor i}}
\State $O_{j}(t) \gets O_{j}(t) + W_{ij} \times \Delta X_{i}(t)$
\State $\Delta X_{j}(t) \gets f(O_{j}(t)) - X_{prev}(t)$

\If{$|\Delta X(t)| \ge T_j$}
	\State $X_{prev}(t) = f(O_{j}(t)) $
	\State \textit{Send $\Delta X_{j}(t)$ to successors}
	
\EndIf
\EndFor
\end{algorithmic}
\end{algorithm}

GrAIMattersLab [15, 16] implemented this algorithm in the NeuronFlow processor architecture \cite{neuronflow_processor, neuronflow_hybrid}.

\subsection{Algorithm}


In the previous sections, we described two neural network optimization algorithms. By combining them, we get the following neural network optimization algorithm:

Stage 1:
\begin{enumerate}
\item Train neural network in the environment
\item Prune top r \% weights
\item Reset weigths to original
\item Repeat steps 1 and 2 n times
\end{enumerate}

Using this algorithm, we obtain a set of structurally sparse neural networks with different degrees of sparsity. The number of neural networks in the set equals the number of pruning algorithm iterations.

Stage 2:
Then we apply the delta network algorithm to these neural networks. As a result, we get a set of new neural networks using both structured and temporal sparsity.
Selection of one neural network from such set depends on the desirable balance between the number of significant multiplications and neural network performance.

\subsection{Significant operations counting}

\subsubsection{Number of multiplications in an unoptimized neural network}

Let us estimate the number of multiplications in a standard neural network without any optimizations. The following formula can be used to count the number of multiplications in convolutional layer k:

\begin{equation}
\# multiplications_{k} = size\_x_{k+1} * size\_y_{k+1} * filters_{k} * kernel\_size\_x_{k} * kernel\_size\_y_{k} * filters_{k},
\end{equation}

where

\begin{itemize}
\item $size\_x_{k+1}$ - is the k+1 layer input size along the x axis
\item $size\_y_{k+1}$ - is the k+1 layer input size along the y axis
\item $filters_{k+1}$ - is the number of filters at layer k (k+1 layer input size along the z axis)
\item $kernel\_size\_x_{k}$ - is the convolution size along the x axis
\item $kernel\_size\_y_{k}$ - is the convolution size along the y axis
\item $filters_{k}$ - is the k layer input size along the z axis
\end{itemize}

The following formula can be used to count multiplications in dense layer $k$:
\begin{equation}
\# multiplications = input_{k} * output_{k},
\end{equation}
where

\begin{itemize}
\item $input_{k}$ - is the k layer input size (number of neurons at k-1 layer) 
\item $output_{k}$ - is the k layer output size (number of neurons at it)
\end{itemize}

General results for all layers are presented in table \ref{table: ops_common}. It should be noted that these results are universal for any game and for any neural network inference.

\begin{table}
\begin{tabular}{ |c|c|c|c| } 
\hline
Layer &  Multiplications  & Param  \\
\hline
Conv2d-1 & 3,276,800 & 8,224 \\
Conv2d-2 & 2,654,208 & 32,832 \\
Conv2d-4 & 1,806,336 & 36,928 \\
Flatten & 0 & 0 \\
Dense-1 &  1,605,632 & 1,606,144 \\
Dense-2 & $512 * n_{output}$ & $512 * n_{output} + n_{output}$ \\
\hline
\end{tabular}
\caption{Number of multiplications in an unoptimized neural network}
\label{table:ops_common}
\end{table}

\subsubsection{The number of multiplications in optimized neural networks}

It is clear that the degree of weight sparsity will affect the number of non-zero multiplications. It should be taken into account that the percentage of pruned weights can differ in different layers.

Using the delta layer provides different levels of temporal sparsity depending on selected threshold, layer, and input data. That is why average statistics on neuron activations was used for neural network runs. Examples of estimated numbers of significant multiplications are given in the next section and in tables 
 \ref{table:ops_breakout}, \ref{table:ops_spaceinvaders}.

\section{Results}

\begin{figure}
\hfill
\subfigure[Breakout]{\includegraphics[width=1.0\textwidth]{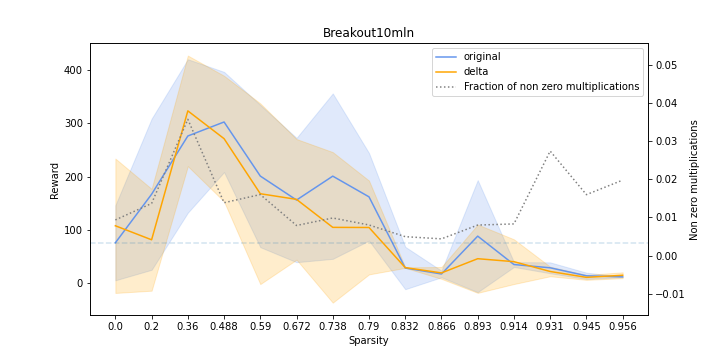}}
\hfill
\subfigure[Enduro]{\includegraphics[width=1.0\textwidth]{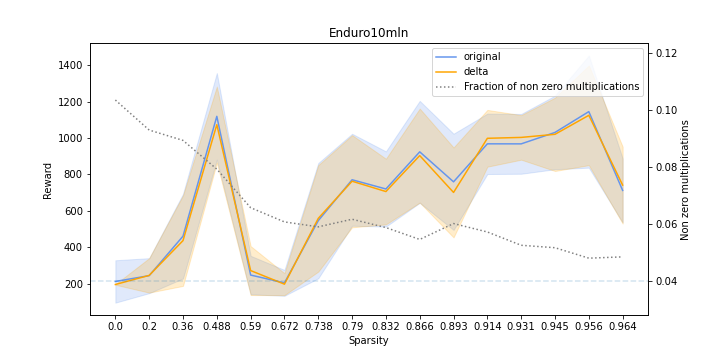}}
\hfill
\caption{Results for SpaceInvaders and Breakout. The x axes of the figures denote the neural network sparsity degree; the left y axes denote the reward received by an agent; the right y axes denote fraction of significant multiplications averaged by environment runs. The orange line shows the performance of the pruned network, while the blue line shows the performance of the pruned network with the additional application of the DeltaNetwork algorithm. The grey dotted lines show the fraction of significant multiplications (the less the better) of a neural network pruned using the DeltaNetwork algorithm. The blue dashed lines demonstrate the performance of a neural network without any optimizations.}
\label{fig:results}
\end{figure}

\begin{figure}
\subfigure[Robotank]{\includegraphics[width=0.5\textwidth]{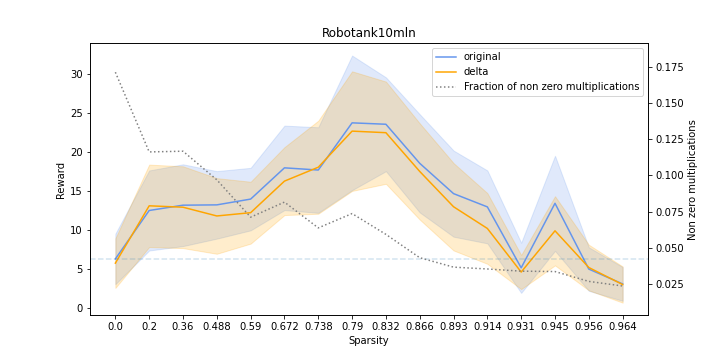}}
\subfigure[Freeway]{\includegraphics[width=0.5\textwidth]{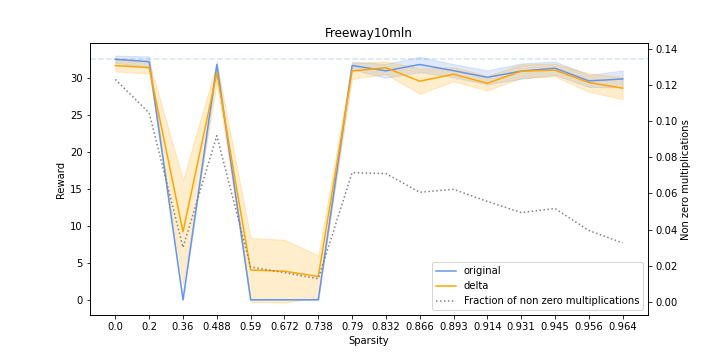}}
\subfigure[BattleZone]{\includegraphics[width=0.5\textwidth]{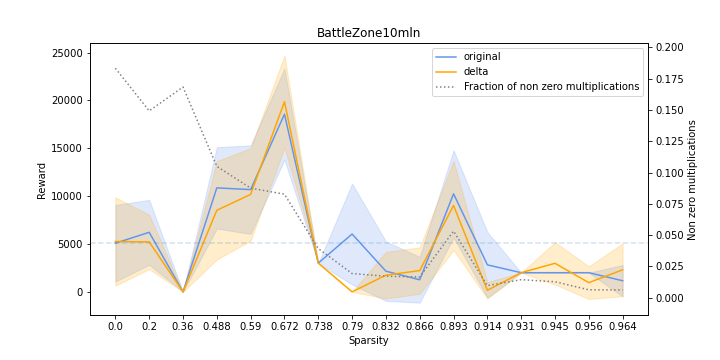}}
\subfigure[Spaceinvaders]{\includegraphics[width=0.5\textwidth]{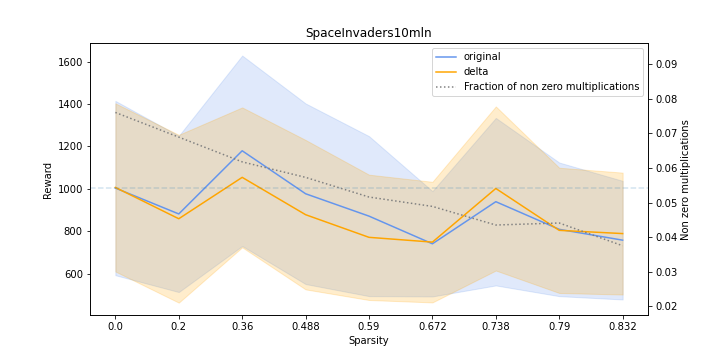}}
\caption{Results for Robotank, BattleZone, Enduro and Freeway.}
\label{fig:results2}
\end{figure}

Figures \ref{fig:results}, \ref{fig:results2} shows the abovementioned neural network performance metrics at different sparsity levels and for different environments.

80 \% of convolutional weights can be pruned for Breakout; notably, the in-game performance will be better than that of an unoptimized neural network version. The results of estimating the number of significant multiplications at this sparsity level and with the delta algorithm threshold of 0.001 are shown in  \ref{table:ops_breakout}. As we can see, the total number of non-zero multiplications is  $75012$, which is $124$ times less than the number of multiplications in an unoptimized neural network.

For Enduro 96 \% of weights can be pruned without the loss of game performance (it is even higher). However, in comparison with Breakout, the efficiency gain is only about $21$ times. 

For the Robotank environment it is also possible to get $30$ times efficiency gain without losing game performance.


Results for BattleZone, SpaceInvaders and Freeway are not as good as in previous environments. 

For example, in SpaceInvaders the delta neural network variant allows playing without losing performance at 73 \% sparsity and with $406486$ significant multiplications, which is $22.9$ times less than the number of multiplications in an unoptimized neural network. A layer-wise operation analysis is shown in table \ref{table:ops_spaceinvaders}.

In BattleZone and Freeway performance plots we can find the points in which the performance is equal to or even better than the performance of the original agent. However, we found an interesting phenomenon. There were pruning levels at which we weren't able to train our agents (zero performance on plots). Unfortunately, we are not able to explain this behavior. It will be done in future research.



The results turn out to be dependent on the game the agent is playing. This can be explained by the fact that the SpaceInvaders game has a lot more changing pixels at each time step than the Breakout game (in the Breakout only the playground and the ball move, while in the SpaceInvaders several objects can move at once - shots, a ship and aliens). This is well confirmed by the difference in Delta sparsity level when playing Breakout and SpaceInvaders (see tables \ref{table:ops_breakout} \ref{table:ops_spaceinvaders}).

Our reward metric results are similar to the results from \cite{lotteryRL1, vischer2021lottery}, where the worst results were in the SpaceInvaders with performance dropping very quickly as neural network sparsity grew.

\begin{table}
\begin{tabular}{ |c|c|c|c|c| } 
\hline
Layer & Multiplications & Nonzero multiplications & Sparsity weights & Delta sparsity  \\
\hline
Input & 0 & 0 & 0.0 & 0.992 \\
Conv2d-1 & 3,276,800 & 9468 & 0.638 & 0.99 \\
Conv2d-2 & 2,654,208 & 5592 & 0.789 & 0.968 \\
Conv2d-4 & 1,806,336 & 10125 & 0.824 & 0.969 \\
Dense-1 & 1605632 & 49790 & 0.0 & 0.975 \\
Dense-2 & 2,048 & 51 & 0.0 & 0.871 \\
\hline
Total & 9344832  & 75012 & 0.79 & 0.987 \\
\hline
\end{tabular}
\caption{Number of multiplications in Breakout with 0.79 sparsity and 0.001 threshold }
\label{table:ops_breakout}
\end{table}

\begin{table}
\begin{tabular}{ |c|c|c|c|c| } 
\hline
Layer & Multiplications & Nonzero multiplications & Sparsity weights & Delta sparsity  \\
\hline
Input & 0 & 0 & 0.0 & 0.986 \\
Conv2d-1 & 3,276,800 & 16712 & 0.635 & 0.924 \\
Conv2d-2 & 2,654,208 & 58216 & 0.711 & 0.773 \\
Conv2d-4 & 1,806,336 & 88558 & 0.784 & 0.849 \\
Dense-1 & 1,605,632 & 242450 & 0.0 & 0.732 \\
Dense-2 & 2,048 & 548 & 0.0 & 0.004 \\
\hline
Total & 9344832  & 406486 & 0.737 & 0.936 \\
\hline
\end{tabular}
\caption{Number of multiplications in SpaceInvaders with 0.74 sparsity and 0.001 threshold}
\label{table:ops_spaceinvaders}
\end{table}



\subsection{Hardware Problem}

Despite clear advantages of this approach, there are very few opportunities to use the existing hardware to effectively implement it today. This is due to modern GPUs being designed for handling dense matrices. Nevertheless, there are attempts to turn the situation.
Nvidia began offering hardware support of sparse matrix operations on one of its latest Tesla A100 GPUs; however, the maximum supported sparsity is only 75 \% so far 
 \cite{krashinsky2020nvidia}.

The authors of the abovementioned DeltaNetwork algorithm work for GrAIMatterLabs that introduced the NeuronFlow architecture-based GrAIOne processor that supports delta neuron based neural network design.

The Loihi2 processor that Intel presented \cite{loihi2} in September 2021 also supports the multi-core asynchronous architecture and running sparse delta neuron-based neural networks.

\section{Conclusion}
This study is the first to demonstrate large multiplication redundancy in inference of neural networks handling RL tasks. Minimizing the number of multiplications becomes critical in areas where computation energy efficiency is important. Such areas include Edge AI and robotics. When speaking of the latter, RL algorithm optimization becomes extremely relevant.
Although there is currently no suitable hardware available to take full advantage of these benefits, this research highlights the importance and potential of this area.

\newpage
\printbibliography

\end{document}